\documentclass[a4paper, 10 pt, conference]{ieeeconf}  

\IEEEoverridecommandlockouts                              

\usepackage{amsmath} 
\usepackage{amssymb}  
\usepackage{graphicx}
\usepackage{float}
\usepackage[T1]{fontenc}
\usepackage[rgb]{xcolor}
\usepackage[author=LSB]{pdfcomment}

\usepackage{multirow}
\usepackage{arydshln}

\title{\LARGE \bf
Implementation of haptic communication in comanipulative tasks: a statistical state machine model 
}

\author{Lucas Roche$^{1}$ and Ludovic Saint-Bauzel$^{1}$
\thanks{$^{1}$Institut des Systèmes Intelligents et de Robotique,
        Université Pierre et Marie Curie, 75005 Paris, France
        {\tt\small roche@isir.upmc.fr}
        {\tt\small saintbauzel@isir.upmc.fr}}
}

\begin{document}

\maketitle
\thispagestyle{empty}
\pagestyle{empty}

\begin{abstract}

This paper presents an experimental evaluation of physical human-human interaction in lightweight condition using a one degree of freedom robotized setup. It explores possible origins of Physical Human-Human communication,  
more precisely, the hypothesis of a time based communication. To explore if the communication is correlated to time a statistical state machine model based on physical Human-Human interaction is proposed. The model is tested with 14 subjects and presents results that are close to human-human performances.  

\end{abstract}

\section{INTRODUCTION}

\begin{figure*}
\centering
  \includegraphics[width=0.95\textwidth]{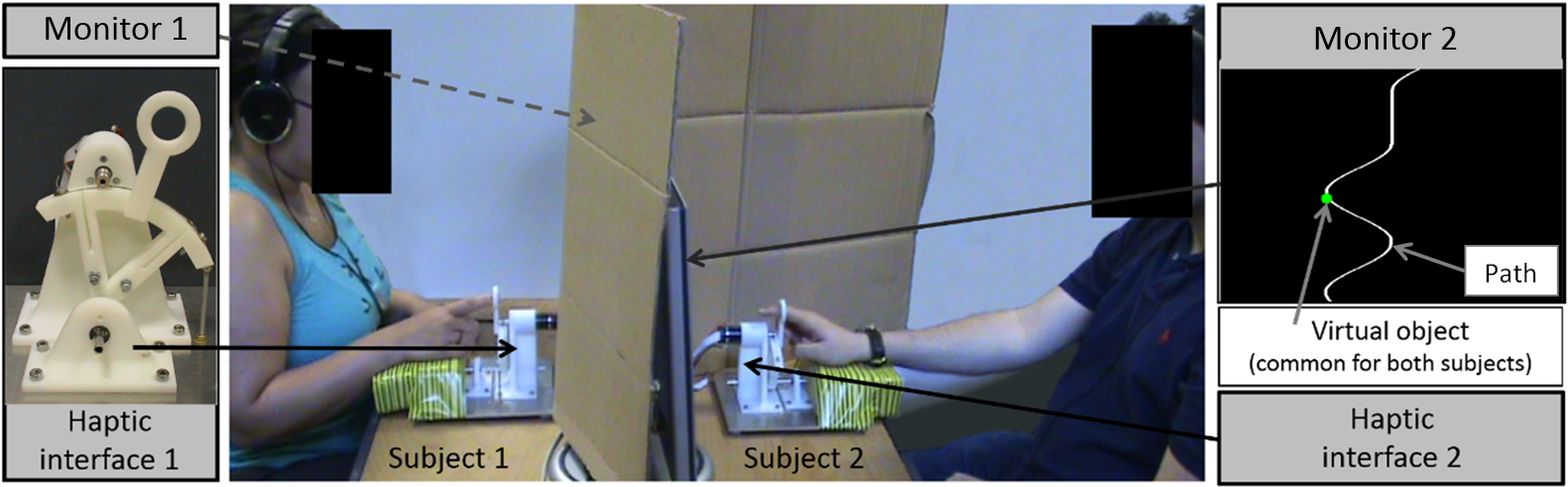}
  \caption{Description of the experimental setup: The two subjects use a one dof haptic interface to share the control over a virtual object. Visual feedback about the position of the object is given on their respective screen as a cursor.}
  \label{fig:setup}		
  \vspace{-0.5cm}
\end{figure*}

Robotic devices have progressed from fully autonomous robots to systems that can share their actions with humans.
Due to the dangerous nature of rigid high-velocity robots, the environment used to be constrained, and  human presence restricted during operation. But recent developments and proofs of safety have brought back the human in the robot workspace. Robots are now more often bound to work alongside humans and to cooperate with them to accomplish a wide range of tasks\cite{peshkin2001}. This cooperation often implies co-manipulation, and hence interaction via direct -- or indirect --  physical contact. 

This contact between humans and robots brings many issues into the design of robots. Indeed, the complexity of human behavior makes us unpredictable for machines.
Peshkin et al. have proposed numerous robot designs called cobots\cite{peshkin2001}, which  are currently able to work alongside the human operator.
The issues of human-robot contact are there addressed using various impedance control methods \cite{santis2007}. Impedance control allows disturbance rejection and uncertainty management, and has been proven extremely efficient, from industry to medical robots.  

In physical human-human co-manipulation, the role distribution amongst the partners is important to achieve the task. 
Jarassé et al.\cite{jarasse2013} explain that multiple role combinations can be used efficiently. According to their framework, a co-manipulative task can be defined as a cooperative task in which a follower/leader approach is an optimal strategy, but no hypothesis on which partner has to be the leader is made. 
Using pure impedance control generally means imposing a master/slave behavior for the human (master) and the robotic partner (slave). While this is sufficient to ensure good performances in most co-manipulative tasks, it struggles to reach the performances obtained in human-human co-manipulation. 
Our assumption is that the fixed role distribution in human-robot interaction cannot compare to the efficiency of flexible roles in human-human interaction. The ability to change roles, that can be observed in human-human cooperation, implies the existence of some kind of communication, or common understanding, during the operation.

The hypothesis that there is communication in haptic co-manipulation has already been explored. 
Multiple studies have been focusing on physical Human-Human Interaction (pHHI), as well as physical Human-Robot Interaction (pHRI). The objective of studying the way humans cooperate in co-manipulation situations is to be able to replicate these behaviors, or at least understand them well enough to help robots adapt to humans.
Reed et al. \cite{reed2008} demonstrated that in one degree-of-freedom pointing tasks, humans dyads performed better than individuals, result confirmed in several other studies \cite{glynn2000} \cite{feth2013} \cite{melendez2011} \cite{madan2014}. They also observed that replicating human behavior could be easily done and can reduce the perceived difference between a human partner and a robotic partner. However, a simple reproduction of the human behavior on a task doesn't seem sufficient to reach the same performances in pHRI than in pHHI \cite{avraham2012} \cite{evrard2009} \cite{ganesh2014}. 
The work of Ganesh \& al. \cite{ganesh2014} further confirmed that pHHI allows to enhance performances, even when the human subject isn't aware of the presence of a partner. Still, neither a perfectly operating partner, nor a recording of a previously performing human can reach the performances (both in terms of precision and improvement), that a pair of humans are able of.
 
These experiments show that there is an advantage in collaboration, which leads to suppose the existence of some kind of haptic communication. 
Humans seem to be able, somehow, to communicate through physical interaction - either consciously or not. 
The existence of this communication has been more precisely observed by Groten\&Feth \cite{feth2013}: in their experiment, they first obtained results comparable to Reed \cite{reed2008}, this time for tracking tasks; secondly, they compared performances of human dyads in a co-manipulative tracking task, depending on the presence -- or not -- of haptic feedback between the subjects. Results show that humans indeed perform better when they are able to feel the action of their partner through haptics, confirming a possibility of communication via haptic channel. 

If Groten \& Feth \cite{feth2013} demonstrated that haptic channel enable  existence of communication, the following questions remains: How does it work? Can we model this communication? This paper will try to explore some hypothesis on the origin of the haptic communication in co-manipulative tasks. The proposed methodology is to analyse experimental data and extract the feature that is the most able to predict the choice of the dyad. 

To explore these question, a specific haptic teleoperated robot will be used, its control and the statistical state model is presented in (\ref{sec:MatMet}) Material and Method section. Then the experimentation protocol and the assessment method is explained in a (\ref{sec:ExpPro}) Experimental protocols section. Explored questions are presented and discussed in (\ref{sec:Res}) Results.

\section{MATERIAL AND METHODS}
\label{sec:MatMet}

The present work is dedicated to the study of the communication through haptics in co-manipulative precision tasks, namely, tasks involving low amplitude (hand/fingers movements) and low efforts (less than 5 Newtons).

%



\subsection{Material}
\label{subsec:mat}


The system is constituted of two one-degree-of-freedom handles.
The mechanical design of each handle, called haptic interface, is inspired from the Stanford University's HapKit \cite{hapkit2014} and can be seen in Figure \ref{fig:setup}. The actuation is done by a DC motor connected to the handle through a cable transmission. The user places his finger on the handle, and can perform leftward or rightward motions during the task.

The controller of the handles and the data acquisition is implemented on a Real-Time operating system (Xenomai - 1 kHz actualization frequency), while the graphical interface runs on another computer. The communication between the two computers is realized by a direct Ethernet connection and the use of UDP and TCP protocols. The average time-delay in this connection is 0.2 ms and is deemed negligible compared to human response time.

\begin{figure}
  \includegraphics[width=\linewidth]{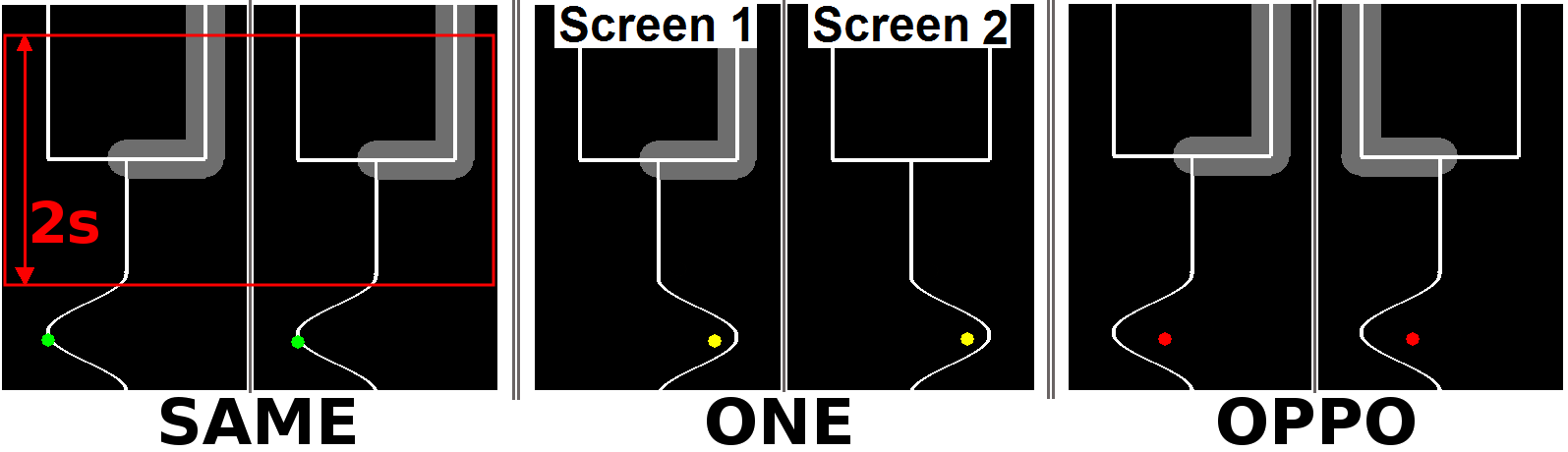}
  \caption{Illustration of the different decision types: SAME, ONE and OPPO.
The data about the choices is recorded from a 2s timezone around the path's fork (in red on the leftward figure).  
Visual feedback about the dyad's performance is given through the color of the cursor (from left to right : green, yellow, red).}
  \label{fig:decisiontypes}
  \vspace{-0.3cm}
\end{figure}

The control of the two handles is based on a teleoperation scheme with a virtual inertia added. They share a position-based control over a virtual object: the position of the virtual object is equal to the mean position of the two handles (each handle thus contributes to half of the displacement/velocity of the virtual object). The position of the virtual object is displayed for each subject on their respective monitor as a cursor (see Figure \ref{fig:setup}). This cursor can move horizontally depending on the position of the handles. 
The co-manipulative task that the subjects have to complete is a tracking task: a path (white line over black background
) is scrolling down on their monitor, at a speed of 35mm/s. The subjects are asked to keep the position of the common virtual object (cursor) as close as possible to the scrolling path.
To further incite each subject to cooperate, they are told that their goal is to maximize the common performance of the dyad. Feedback about the common performance is given by the color of the cursor, which changes based on the distance between the closest path and the cursor (see Figure \ref{fig:decisiontypes}):
\begin{itemize}
\item Green if $ \left| X_{cursor} - X_{Path} \right| < 5 mm$
\item Orange if $ 5mm < \left| X_{cursor} - X_{Path} \right| < 10 mm$
\item Red if $ \left| X_{cursor} - X_{Path} \right| > 15 mm$
\end{itemize}

The path is composed of a semi-randomly generated succession of curves, divided in two categories. 
\begin{enumerate}
\item The "BODY" category, is composed of sinusoidal paths of random direction but fixed duration. The purpose of these parts is to keep the subjects focused on the task between two of the studied parts.
\item The "CHOICE" category is the aim of the experiment: at fixed intervals, the path splits into a fork, imposing a clear choice to be made concerning the direction that the subjects need to follow (see Figure \ref{fig:decisiontypes}). Considering that the subjects can neither see nor hear each other, the only way they can come to an agreement about the direction to choose is to use either the visual feedback of the monitor, or the haptic feedback of the handles. 
\end{enumerate}

While the path's structure is strictly the same for both subjects, each subject is encouraged to follow an highlighted trajectory. During the CHOICE parts, subjects receive some information about what side they have to choose \cite{feth2013}; the given information can differ, which artificially creates situations of agreement or conflict, distributed in three cases. This is done by highlighting one of the two paths of the fork (see Figure \ref{fig:decisiontypes}):
\begin{itemize}
\item \textbf{SAME}: Both subjects have the same information, no conflict occurring.
\item \textbf{OPPO}: Opposite information given to each subject, inducing a conflicting situation.
\item \textbf{ONE}: Only one subject has the information, this condition forces the subjects to be ready to take initiative in case they are the only one having information about the path to choose. ONE exists to discourage subject to keep a passive strategy all along the trial. 
\end{itemize}
The subjects are informed about these choices and the different decision types beforehand.

\subsection{CHOICE Parts analysis}

Each Choice Phase is composed of a straight line lasting one second, followed by a fork where the path splits into two different paths. The paths merge again after 3 seconds of straight line.
The analysis is conducted over a 2 seconds duration around the fork (see Figure \ref{fig:decisiontypes}).
The horizontal position of the cursor is noted $X_{cursor}$. A negative value of $X_{cursor}$ means that the cursor is on the left, a positive value means that the cursor is on the right.
After the fork, the leftward and rightward paths are respectively situated at $X_{left} = -X_{MAX}$ and $X_{right} = X_{MAX}$,\\ with $X_{MAX} = 80\ pixels \simeq 25 mm$.


\subsection{Intention predictor}
\label{subsec:intentpred}

%

We define a predictor as a physical parameter which can be used to predict the direction that a dyad will take in a CHOICE part.
In order to find the parameter that best predicts the behavior of the dyad, an analysis of experimental data is performed. The data was obtained with seven pairs of subjects using the setup.

The optimal predictor should have a perfect accuracy, and obtain results as early as possible during the CHOICE part, in order to allow for reaction time from both the robot and the human.
Analysis of the experimental data shows that most predictors increase in accuracy as the analysis time approaches the fork ($t_{choice} = 1s$).

However, the analysis must end before completion of the actual motion. In order to respect this constraint, the analysis end time is fixed at 0.9 second after the start of the CHOICE part. This timing is based on the data analysis: only 5\% of the motions are completed at 0.9 second, which is considered an acceptable margin of error.

Many predicting parameters can be defined and compared:
\begin{itemize}
\item XT : Position of the cursor at time $t_{stop}$\footnotemark[1]. 
\linebreak $X_T = X_{cursor}(t_{stop})$
\item XM : Mean position over $[t_{start};t_{stop}]$\footnotemark[1]. 
\linebreak $X_M = \sum\limits_{k=1}^N{\frac{X_{cursor}(k)}{N}}$
\item VT : Instantaneous velocity at time $t_{stop}$\footnotemark[1]. 
\linebreak $V_T = \dot{X}_{cursor}(t_{stop})$
\item VM : Mean velocity over $[t_{start}; t_{stop}]$\footnotemark[1]. 
\linebreak $V_M = \sum\limits_{k=1}^N{\frac{\dot{X}_{cursor}(k)}{N}}$
\item FM : Mean sum of forces applied on the handle over $[t_{start}; t_{stop}]$\footnotemark[1]. 
\linebreak $F_M = \sum\limits_{k=1}^N{\frac{F_{Subject 1}(k) + F_{Subject 2}(k)}{N}}$
\item Mean signed RMS deviation. 
\linebreak $SRMS = \sum\limits_{k=1}^N{\frac{ (X_{cursor}(k)-X_M)*\left| X_{cursor}(k)-X_M \right|}{N}}$ 
\item 1C : First Crossing of a threshold ($X_{TH}$) :
	\linebreak \emph{$if(X_{cursor}(t)>X_{TH}) \to 1C=right$ \linebreak $ if(X_{cursor}(t)<-X_{TH}) \to 1C= left$}.
\end{itemize}

\begin{table}
\label{table:predictors}
\caption{Accuracy of the different predictors at 0.9 second after the start of the CHOICE Phase (* for this criteria 0.9 second is the average crossing time)}
\begin{center}
\begin{tabular}{|c|c|}
\hline 
Predictor & Accuracy at 0.9 second (\%) \\ 
\hline
XT & 87.05 \\ 
\hline 
VM & 86.81 \\ 
\hline 
VT & 86.36 \\ 
\hline 
FM & 82.72 \\ 
\hline 
XM & 74.09 \\ 
\hline 
SRMS & 53.43 \\ 
\hdashline 
\textbf{1C*} & \textbf{94.23} \\
\hline
\end{tabular}
\end{center}
\vspace{-0.5cm}
\end{table}

\footnotetext[1]{$t_{start}$ and $t_{stop}$ are variable parameters allowing to tune the size of the interval in which data is analysed. The results presented here are obtained with values of these parameters which maximize the precision.}

The precision of these predictors at 0.9 second is exposed in table \ref{table:predictors}. As we can see, the only predictor that is able to reach over 90\% accuracy is the First Crossing parameter.

The First crossing parameter (1C) is constructed based on the hypothesis that initiative of the subjects is a preponderant factor in the choice of the dyad. This parameter is  designed experimentally.

\begin{figure}[!bp]
\centering
  \includegraphics[width=0.9\linewidth]{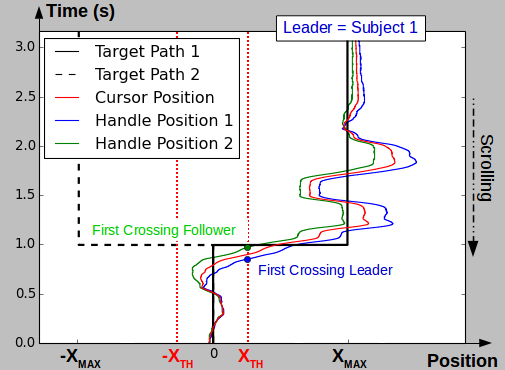}
  \caption{Description of the First Crossing parameter: time at which the individual position of a subject drift from the central position of 30\% of the total target distance. Results show that the Leading subjects have statistically lower starting times.}
  \label{fig:stTime}
\end{figure}

We define the First Crossing parameter as the time at which the individual position of one of the two subjects exits the interval $[-X_{TH}; X_{TH}]$. An illustration can be seen on Figure \ref{fig:stTime}.

The accuracy of the First Crossing predictor (1C) is critically linked to the time needed to reach the threshold. Indeed, if increasing the threshold size enhances the performance, it also increases the time at which the threshold is crossed, and thus the time at which the analysis is completed.
Considering the strong time constraint over the prediction, it is mandatory to select parameters allowing to obtain an average crossing time inferior to 0.9 second. At the same time, the predictor's accuracy needs to the greatest possible.
Choosing a value of $X_{TH}$ equal to 30\% of the total distance between the middle path and one side path allows to reach an accuracy of 94.23\%, with an average analysis time of 0.899 second, which satisfy all of our constraints.
The First Crossing time detected is always at least 0.1 second before the motion ending time, with an average of 0.20 second between the two, which is sufficient for the robotic system to react.

\subsection{Statistical state machine model}
\label{subsec:statemach}

These findings are used to design an algorithm which can reproduce the observed behavior, while staying as simple as possible. The objective is to evaluate how this algorithm can perform as a partner in a cooperative precision task.

\begin{figure}
\centering
  \includegraphics[width=0.9\linewidth]{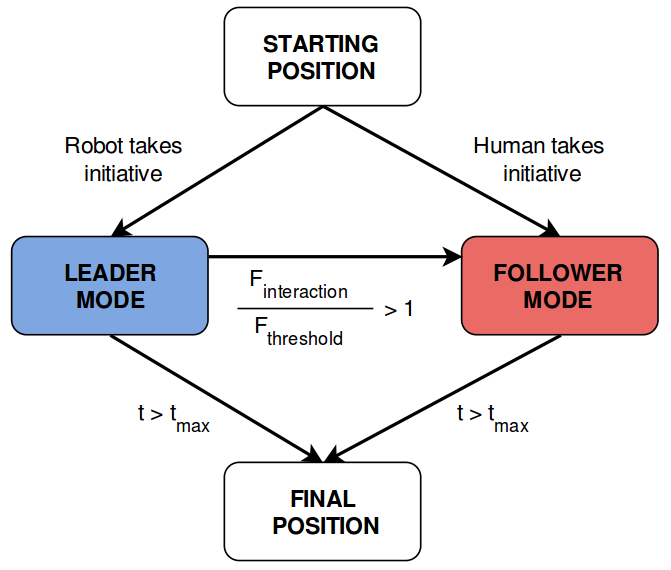}
  \caption{Schematic functioning of the algorithm. The algorithm is designed to let the human lead the movement as a default. In the absence of human initiative, the robot engage the movement toward its own target.}
  \label{fig:algo}
  \vspace{-0.5cm}
\end{figure}
The behavior of the algorithm (see Figure \ref{fig:algo}) is designed to model human.
Therefore the algorithm only has access to information that would be otherwise available to a human subject: (a) The target trajectory; (b) The position of its handle (simulated); (c) The position of the cursor on the monitor; (d) The effort transmitted through the handle. Indirectly, the algorithm can also determine the position of its partner's handle (through the position of the cursor and its own handle).

In the BODY parts, the algorithm follow the path. The controller parameters are adjusted in these parts to induce a lower stiffness, in order to avoid a "too perfect to be human" feeling from the human user.
When confronted to a CHOICE, the algorithm generates a minimum-jerk trajectory\cite{flash_coordination_1985} from its current position to the target position, based on the choice it has to make. The starting time for this trajectory is generated from a normally distributed variable based on the average and standard deviation of the human behavior data.
(In a ONE decision type trial where the robot doesn't have a privileged choice, a direction is chosen at random, with a greater starting time).

If the human partner initiates\footnotemark a motion before the starting time of the robot, the robot lets the human lead, entering "Follower mode". If the human started a motion in the direction the robot wanted to head, nothing changes, otherwise, the robot generates a new trajectory to follow.
If the human partner did not initiate a motion before the beginning of the robot's trajectory, the robot enters "Leader Mode" and initiates the movement along its planned trajectory.
\footnotetext{Taking the initiative is here defined as engaging a movement of the handle resulting in a displacement of the virtual object superior to 30\% of the distance between the starting position and the target.}

The robot can leave the "leader mode" if the interaction force exceed the force threshold ($F_{threshold} = 0.7 N$) for a duration of $\Delta T_{th} = 0.2 s$.

While in "Follower Mode", the robot can still change its trajectory if the interaction force condition is met again, and change its target between left and right to follow the human.

\section{Experimental protocols}
\label{sec:ExpPro}

The 2013 study of Feth \& Groten proves the existence of human-human communication through the haptic channel, although their study is restricted to higher load co-manipulative tasks. This experimental protocol is here applied to a light weighted setup, the aim being to study the possibility to predict the behavior of a human dyad. This protocol will be summed-up in the following section and the criteria are also re-explained here. However to have a more detailed description, one can refer to \cite{feth2013}.  
  

The experiment is composed of many trials involving each time a pair of subjects. A trial is defined as two human subjects realizing a co-manipulative tracking task as presented in Figure \ref{fig:setup}. 

The three possible decision types (SAME, ONE, OPPO) presented in \ref{subsec:mat}, are combined with three experimental conditions:
\begin{itemize}

\item \textbf{Haptic-Feedback-from-Object-and-Partner (HFOP)}: the positions of the handles are kept identical by a PID controller with high gain simulating a rigid connection\cite{niemeyer1991}.
In addition, the virtual object is given a simulated inertia of 40 grams. These 40 grams represent the weight of a common surgical tool.  The visual feedback of the cursor is identical for both subject and represents the position of the virtual object.

\item \textbf{Hidden Robotic Partner (HRP)}: the subjects believe they are doing the task together, while they are actually independently performing their task each paired with their own robotic partner. The subjects have visual feedback concerning their own task and virtual object on their monitor, and can feel the haptic feedback from the virtual inertia and the efforts of their robotic partner.


\item \textbf{Subjects separated (ALONE)}: the two subjects perform the same tracking task independently. Two separated setup constituted of one of the haptic interfaces and one monitor are used for each subject. Two different virtual objects are simulated and each subject executes the task alone, with visual feedback from the monitor, and haptic feedback from their own virtual object.
This experimental condition serves two principal purposes: (1) It draws a comparison between the performances of a single subject with the performances of a dyad.
(2) It is used as a buffer trial between the other 2 conditions, to reduce the potential learning effects and sensory memory between HFOP and HRP trials.
\end{itemize}


Each experiment is a succession of 6 trials, which can be ordered in two different configurations a and b.


\begin{center}

\begingroup\makeatletter\def\f@size{5}\check@mathfonts
\def\maketag@@@#1{\hbox{\m@th\large\normalfont#1}}%

\hspace{-1,5\parindent}\begin{tabular}{c|c|c|c|c|}
\cline{2-5}
a)&$HRP(\times2)$ & \multirow{2}{1,3cm}{$ALONE(\times2)$}  & $HFOP (\times2)$  \\\cdashline{2-2}\cdashline{4-4}
b)&$HFOP(\times2)$ &                   & $HRP (\times2)$  \\
\cline{2-5}
\end{tabular}
\endgroup
\linebreak
\end{center}

At the beginning of the experiment, the subjects are explained the rationale of the setup and told about the different choices in the task. They are also told that two different experimental conditions are tested: they can either perform the task alone (ALONE), or cooperate through comanipulation (HFOP). In reality, the HRP condition is also tested, although the subjects are unaware of this fact.
The order between HRP and HFOP in the first and third parts of the experiment is drawn at random beforehand. 
For each pair of subjects, there are two trials per condition; the first one is used as training, the second one for data analysis. Each trial is 120s long, containing 16 choices randomly picked from one of the 3 different decision types (SAME, ONE, OPPO). 

The subjects are physically separated by an opaque screen to prevent any visual clue about the actions of their partners, and wear audio headphones playing white noise to prevent any auditory clue.

\section{RESULTS}
\label{sec:Res}
All comparison between the different statistical data sets presented here are calculated using a Student's t-test.

\subsection{Performance criterion}

We measured the precision of the dyad in the task with a performance criterion \cite{feth2013}. This parameter is obtained by first calculating the RMS:
\begin{equation}
RMS = \sqrt{\frac{\sum\limits_{k=1}^N (x_{t,k} - x_{o,k})^2}{N}}
\end{equation}
where $x_{t,k}$ and $x_{o,k}$ are respectively the target position and the virtual object position at time step k.
This RMS error for a choice is then compared to $RMS_{max}$ the maximum RMS obtained on the whole sample of trials
\begin{equation}
Performance = 1 - \frac{RMS}{RMS_{max}}
\end{equation}
This performance indicator is preferred over RMS error for clarity:  the better the results, the greater the performance.

%
%
%
%
%
%
%

\subsection{Experimental results: Statistical state machine evaluation experiment}
\label{sec:res2}
The experiment was conducted on 14 subjects (mean age 22.9, 11 males, 3 females), for a total of 1120 choices recorded (224 in HFOP, 448 in ALONE, 448 in HRP).


\subsubsection{Change of scale}
The change of scale doesn't affect the general behavior of the subjects during the experiment. The results obtained during the experiments lead to performances that are similar to those found by Feth\&Groten\cite{feth2013}.

\subsubsection{Performances}

\begin{figure}
  \includegraphics[width=\linewidth]{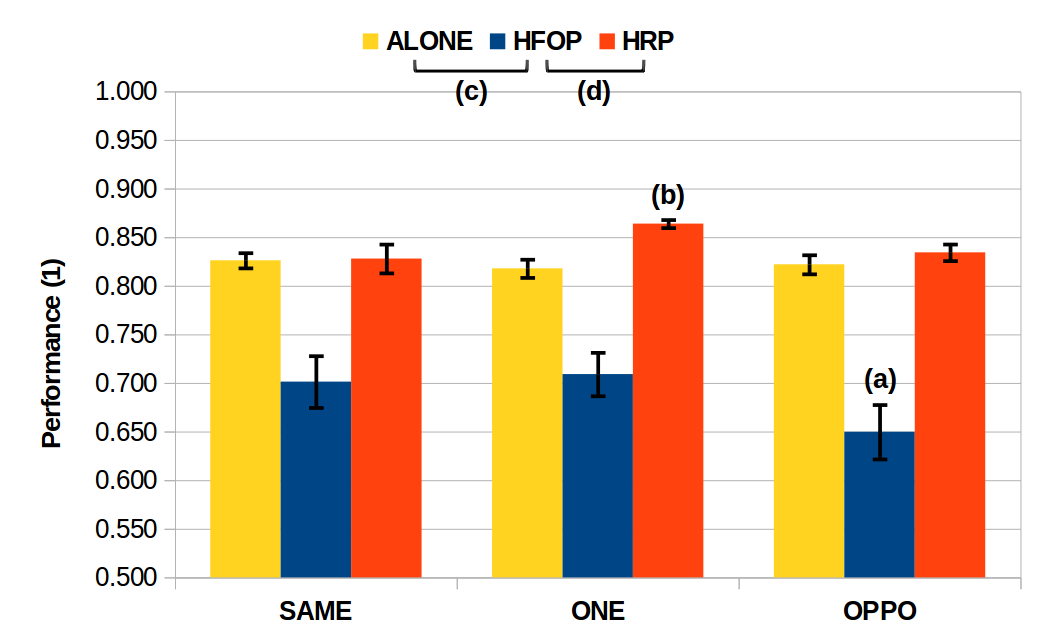}
  \caption{Experiment 2 - Results expressed in terms of Performances, for the three experimental conditions (ALONE, HFOP and HRP) and the three decisions types (SAME, ONE and OPPO)}
  \label{fig:perfs-manip2}
  \vspace{-0.5cm}
\end{figure}

The results of this section are illustrated in Figure \ref{fig:perfs-manip2}.

There is a significant effect of the decision type on performance, both for the HFOP and HRP condition. 
(a) In HFOP, the OPPO decision type leads to significantly worse performances than SAME and ONE ($p<0.005$ for both).
(b) In HRP, ONE is significantly better than both SAME and OPPO ($p<0.005$ for both).
In ALONE, no significant difference is found between the decision types.

The performance is also significantly affected by the experimental condition : 
(c),(d)The HFOP condition leads to worse performances in all three decision types ($p<0.001$); The HRP condition leads to better performances than ALONE condition for the ONE decision type, and worse for the OPPO one ($p<0.005$ for both).
The influence of the decision type on performance is detailed in Table~\ref{table:perfVSdecisionType}\footnotemark[3], and the influence of the experimental condition is detailed in Table~\ref{table:perfVScondition}\footnotemark[3].
\footnotetext[3]{ (*): p-value < 0.05 - (**): p-value < 0.005}
\begin{table}[h]
\caption{Influence of the Decision Type over Performance}
\label{table:perfVSdecisionType}
\vspace{-0.5cm}
\begin{center}
\resizebox{\linewidth}{!}{%
\begin{tabular}{| l | c | c | c |}
\hline
Condition	& SAME vs ONE 	& SAME vs OPPO	&	ONE vs OPPO 	\\
\hline
HFOP		& $SAME \sim ONE$	& $\textbf{SAME} > OPPO ^{**}$	&	$\textbf{ONE} > OPPO^{**}$\\
\hline
HRP			& $SAME < \textbf{ONE} ^{*}$	& $SAME \sim OPPO$	&	$\textbf{ONE} > OPPO ^{**}$\\
\hline
ALONE			& $SAME \sim ONE$	& $SAME \sim OPPO $	&	$ONE \sim OPPO$\\
\hline
\end{tabular}}
\end{center}
\vspace{-0.5cm}
\end{table}

\begin{table}[h]
\caption{Influence of the Experimental Condition over Performance}
\label{table:perfVScondition}
\vspace{-0.5cm}
\begin{center}
\resizebox{\linewidth}{!}{%
\begin{tabular}{| l | c | c | c |}
\hline
Dec. Type	& ALONE vs HFOP	& HRP vs HFOP	& ALONE vs HRP\\
\hline
SAME			& $\textbf{ALN} > HFOP ^{**}$	& $\textbf{HRP} > HFOP ^{**}$	& $ALN \sim HRP $\\
\hline
ONE			& $\textbf{ALN} > HFOP ^{**}$	& $\textbf{HRP} > HFOP ^{**}$	& $ALN<\textbf{HRP} ^{**}$\\
\hline
OPPO			& $\textbf{ALN} > HFOP ^{**}$	& $\textbf{HRP} > HFOP ^{**}$	& $\textbf{ALN} > HRP ^*$ \\
\hline
\end{tabular}}
\end{center}
\vspace{-0.5cm}
\end{table}


\subsubsection{Starting times}
The starting time of the leader is significantly inferior to the starting time of the follower for both HFOP and HRP conditions ($p<0.001$ for both).
The starting time parameter isn't influenced by the experimental condition.

\subsubsection{Dominance}
In the HRP condition, the human subjects take the lead in 66\% of the choices, against 34\% for the robot, this difference is statistically significant ($p<0.005$).


\subsubsection{Robot alone}
When running without a human partner, the robot reaches performances that are significantly better than humans in HFOP conditions ($p < 0.001$), but worse than humans in ALONE condition ($p < 0.001$).


\subsection{Discussion}
\label{sec:Dis}

Concerning the influence of the experimental condition over the performances: 
the HFOP condition leads to worse performances than the ALONE condition when confronted to the CHOICEs parts. This result seems logical since a subject performing the task ALONE doesn't have to handle the resolution of conflicts, nor does he have to deal with the uncertainty caused by the movements of his partner.

Most importantly, the results obtained in the HRP condition (robotic partner) are significantly better than the results in HFOP (human partner). This is a promising result, considering the relative simplicity of the state-machine algorithm used.
Actually, the human-robots dyads from HRP condition even achieve similar results than the human subjects performing the task alone for the SAME decision type, and even better for the ONE decision type. Only when confronted to the OPPO parts does the HRP condition lead to worse performance, but while still being indubitably better than the HFOP condition. Performing the task with our robotic partner seems to be easier, and to simplify the resolution of conflicting situations, compared to a standard pHHI situation.
These results are reliable since the robot was not designed to be a perfect partner. Actually, the robot alone reaches worse performance than humans in ALONE condition.  
The good performances of the human-robot dyads thus only come from an actual complementarity.
On the same token, while the robot was indeed less dominant in the choice of a path, which comes directly from the conception of the algorithm- it was able to win around one third of the conflicts, and was far for being totally passive.

While encouraging, the results presented in this article still need some deepening. The one degree of freedom setup presents a strong simplification of physical human-human interaction. Furthermore the one alternative is also strongly simplifying the wide range of possible haptic communication. There is room for strengtening the conclusions of this study. 

\section*{Conclusion and perspectives}
The analysis of experimental data acquired with our setup revealed that time is an important factor when communicating through haptic channel. Furthermore, the initiative parameter we defined in \ref{subsec:intentpred} allows us to accurately predict the result of the negotiations in pHHI. Based on these observations, we designed an algorithm aiming to reproduce the behavior of a human partner, and used it in a pHRI experiment.

Considering the observed importance of the initiative in physical interaction, we can make the assumption that humans tend to naturally infer the actions of their partner via tactile clues, and adapt their behavior according to the perceived intentions of this partner. 


  Future work will be done concerning the development of a more dominant algorithm, to study the stability for more complex scenarii (greater number of degree of freedom and of choices), and more conflicting situations. Additionally, the influence of the a priori concerning the nature of the partner was not studied in this paper. Lastly, the initiative, even if proved to be an important factor, can't allow a full reproduction of human behavior in co-manipulative task. New parameters will have to be studied and added to our model to achieve a good human-robot synergy.




%

%
%


\bibliography{bibli} 

\begin{thebibliography}{10}

\bibitem{peshkin2001}
M.~Peshkin and J.~E. Colgate, ``Cobot architecture,'' {\em IEEE Tansactions on
  Robotics and Automation, Vol 17, No 4}, 2001.

\bibitem{santis2007}
A.~D. Santis, B.~Siliciano, A.~D. Luca, and A.~Bicchi, ``An atlas of physical
  human-robot interaction,'' {\em Mechanism and Machine Theory}, 2007.

\bibitem{jarasse2013}
N.~Jarasse, V.~Sanguinetti, and E.~Burdet, ``Slaves no longer : review on role
  assignment for human-robot joint motor action,'' {\em Adaptative Behavior},
  2013.

\bibitem{reed2008}
K.~B. Reed and M.~A. Peshkin, ``Physical collaboration of human-human and
  human-robot teams,'' {\em IEEE Transactions on Haptics, Vol 1, No 2}, 2008.

\bibitem{glynn2000}
S.~Glynn and R.~A. Henning, ``Can teams outperform individuals in a simulated
  dynamic control task?,'' in {\em Proceedings of the IEA 2000/HFES 2000
  Congress}, 2000.

\bibitem{feth2013}
R.~Groten, D.~Feth, R.~L. Klatzky, and A.~Peer, ``The role of haptic feedback
  for the integration of intentions in shared task execution,'' {\em IEEE
  Transactions on Haptics, Vol 6, No 1}, 2013.

\bibitem{melendez2011}
A.~Melendez-Calderon, ``Classification of strategies for disturbance
  attenuation in human-human collaborative tasks,'' in {\em 33rd Annual
  International Conference of the IEEE EMBS}, 2011.

\bibitem{madan2014}
C.~E. Madan, A.~Kucukyilmaz, T.~M. Sezgin, and C.~Basdogan, ``Recognition of
  haptic interaction patterns in dyadic joint object manipulation,'' {\em IEEE
  Transaction on Haptics, Vol 8, No 1}, 2014.

\bibitem{avraham2012}
G.~Avraham, I.~Nisky, H.~Fernandes, D.~E. Acuna, K.~P. Kording, and G.~E. Loeb,
  ``Toward perceiving robots as humans : Three handshake models face the
  turing-like handshake test,'' {\em IEEE Transition on Haptics, Vol 5, No 3},
  2012.

\bibitem{evrard2009}
P.~Evrard and A.~Kheddar, ``Homotopy switching model for dyad haptic
  interaction in physical collaborative tasks,'' in {\em Third Joint
  Eurohaptics Conference and Symposium on Haptic Interfaces for Virtual
  Environment and Teleoperated Systems}, 2009.

\bibitem{ganesh2014}
G.~Ganesh, A.~Tagaki, T.~Yoshioka, M.~Kawato, and E.~Burdet, ``Two is better
  than one: Physical interactions improve motor performance in humans,'' {\em
  Nature, Scientific Report 4:3824}, 2014.

\bibitem{hapkit2014}
T.~Morimoto, P.~Blikstein, and A.~Okamura, ``[d81] hapkit: An open-hardware
  haptic device for online education,'' in {\em Haptics Symposium (HAPTICS),
  2014 IEEE}, 2014.

\bibitem{flash_coordination_1985}
T.~Flash and N.~Hogan, ``The coordination of arm movements: an experimentally
  confirmed mathematical model,'' {\em The journal of Neuroscience}, vol.~5,
  no.~7, pp.~1688--1703, 1985.

\bibitem{niemeyer1991}
G.~Niemeyer and J.~J.~E. Slotine, ``Stable adaptive teleoperation,'' {\em IEEE
  Journal of Oceanic Engineering}, vol.~16, no.~1, 1991.

\end{thebibliography}
\bibliographystyle{ieeetr}

\end{document}